\documentclass[letterpaper, 10pt, conference]{ieeeconf}

\usepackage{amsmath,amssymb,amsfonts}
\usepackage{graphicx}
\usepackage{todonotes}
\usepackage{mathtools}
\usepackage{pgfplots} 
\usepackage{multicol}
\usepackage[keeplastbox]{flushend}
\usepackage{tabularx}
\usepackage{multirow}
\usepackage{caption}
\usepackage{array, graphicx}
\usepackage{subfig}

\usepackage[
    backend=biber,
    natbib=true,
    url=false, 
    doi=false,
    eprint=false,
    maxcitenames=1,
    sorting=none,
    isbn=false
]{biblatex}

\graphicspath{ {figures/} }

\begin{document}

\IEEEoverridecommandlockouts %

\title{\LARGE \textbf{Graph Neural Networks and Reinforcement Learning for Behavior Generation in Semantic Environments}\\
}

\author{Patrick Hart$^{1}$ and Alois Knoll$^{2}$%
	\thanks{$^{1}$Patrick Hart is with the fortiss GmbH, An-Institut Technische Universit\"{a}t M\"{u}nchen, Munich, Germany. Email: patrick.hart@tum.de}%
	\thanks{$^{2}$Alois Knoll is with the Chair of Robotics, Artificial Intelligence and Real-time Systems, Technische Universit\"{a}t M\"{u}nchen, Munich, Germany}%
}

\maketitle

\begin{abstract}
Most reinforcement learning approaches used in behavior generation utilize vectorial information as input.
However, this requires the network to have a pre-defined input-size -- in semantic environments this means assuming the maximum number of vehicles.
Additionally, this vectorial representation is not invariant to the order and number of vehicles.
To mitigate the above-stated disadvantages, we propose combining graph neural networks with actor-critic reinforcement learning.
As graph neural networks apply the same network to every vehicle and aggregate incoming edge information, they are invariant to the number and order of vehicles.
This makes them ideal candidates to be used as networks in semantic environments -- environments consisting of objects lists.
Graph neural networks exhibit some other advantages that make them favorable to be used in semantic environments.
The relational information is explicitly given and does not have to be inferred.
Moreover, graph neural networks propagate information through the network and can gather higher-degree information.
We demonstrate our approach using a highway lane-change scenario and compare the performance of graph neural networks to conventional ones.
We show that graph neural networks are capable of handling scenarios with a varying number and order of vehicles during training and application.
\end{abstract}
\section{Introduction}
\label{sec:introduction}
Many reinforcement learning approaches in decision-making for autonomous driving use vectorial representations as inputs -- e.g.\ a list of semantic objects or images.
However, this requires a pre-defined input-size and order when using conventional deep neural networks.
As a consequence -- in semantic simulations -- the maximum number and order of the vehicles have to be defined.

The number and order of vehicles in real-world traffic situations can change rapidly -- as vehicles come into and leave the field of view or vehicles overtake each other.
Thus, each situation requires an assumption of the maximum number of vehicles and also in which order they should be sensed by the vehicle.
Of course, an arbitrary order of the vehicles could be passed to conventional neural networks during training.
However, this would require the conventional neural network to see all possible combinations during training in order to handle this arbitrary order.
On the contrary, graph neural networks (GNNs) are invariant to the number and order of vehicles as they directly operate on graphs.
This makes them ideal candidates to be used as a decision-making entity in autonomous driving.
\begin{figure}[t!]
	\vspace{2mm}
	\centering
	\def\svgwidth{8.5cm}
	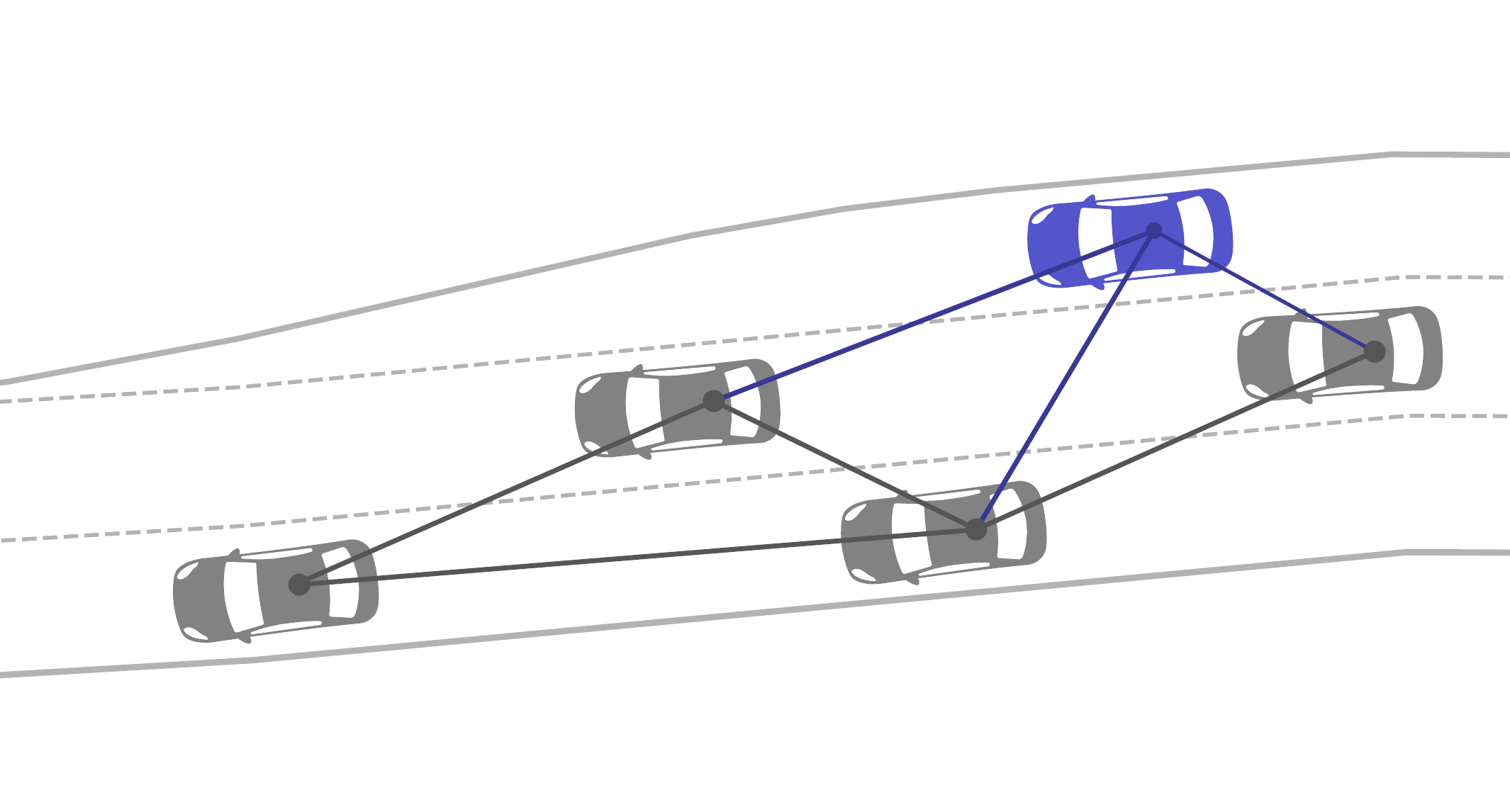
	\caption{Semantic environment in which the vehicles are represented in a graph.
					 The vehicles are nodes and are connected to each other with edges.
					 Graph neural networks take graphs directly as input.}
	\label{fig:intro}
\end{figure}

In this work, we combine continuous actor-critic (AC) reinforcement learning methods with GNNs to enable a number and order invariant decision-making for autonomous vehicles.
AC reinforcement learning methods exhibit \emph{state-of-the-art} performance in various continuous control problems \cite{Morita, Dillmann}.
Additionally to the before-stated advantages, GNNs also introduce a relational bias to the learning problem -- due to the connections between vehicles in the graph.
Thus, relational information is provided explicitly and does not have to be inferred by using collected experiences.
Moreover, GNNs propagate information through the graph due to their convolutional characteristics.
In this work, we use a `GraphObserver' that generates a graph connecting the $n$-nearest vehicles with each other and an `Evaluator' that outputs a reward signal and that determines if an episode is terminal.
Using the `GraphObserver', the `Evaluator' and the AC algorithm, the ego vehicle's policy can now be iteratively evaluated and improved.
The main contributions of this work are:
\begin{itemize}
	\item Using GNNs as networks in AC methods for decision-making in semantic environments,
	\item comparing the performance of conventional deep neural networks to using GNNs and,
	\item performing ablation studies on the invariance towards the number and order of vehicles for both network types.
\end{itemize}

\subsection{Graph Neural Networks}
Graph neural networks (GNNs) are a class of neural networks that operate directly on graph-structured data \cite{Boedecker}.
A wide variety of graph neural network architectures have been proposed \cite{Monfardini, Xing, Cohn, LeCun}.
These range from simple graphs \cite{Monfardini}, to directed graphs \cite{Xing}, to graphs that contain edge information \cite{Cohn}, up to convolutional graphs \cite{LeCun}.

In this work, we use the approach introduced by \citet{Battaglia} that uses a directed graph with edge information.
The graph $G=(N, E)$ is defined having nodes $n_i \in N$ and directed edges $e_{ij} \in E$ from node $n_i$ to $n_j$.
Both -- the nodes and edges -- contain additional information.
The node value is denoted as $\underline{h}_i$ for the $i$-th node and the edge value as $\underline{e}_{ij}$ connecting the $i$-th with the $j$-th node.
The node value $\underline{h}_i$ contains e.g.\ the vehicle's state and the edge value $\underline{e}_{ij}$ relational information between two nodes.
In each layer $k$ of the GNN, a dense node neural network layer is applied per node and a dense edge neural network layer per edge.
Each GNN layer has three computation steps:
First, the next edge values $\underline{e}^{k+1}_{ij}$ are computed using the current edge values $\underline{e}^k_{ij}$, the from-node values $\underline{h}^k_i$  and the to-node values $\underline{h}^k_j$.
These values are concatenated and passed into a (dense) neural network layer $f^k_\chi(\cdot)$ that is parameterized by $\chi$.
This can be expressed as
\begin{equation}
\underline{e}^{k+1}_{ij} = f_{\chi}^k([\underline{h}^k_i, \underline{e}^k_{ij}, \underline{h}^k_j]).
\end{equation}

Next, all incoming edge values $\underline{e}^{k+1}_{ij}$ to the node $n_j$ are aggregated.
In this work, we use a sum as the aggregation function.
Thus, the node-wise aggregation of the edge values can be written as
\begin{equation}
\underline{e}_{agg, j}^{k+1} = \sum_{i=0}^{M} \underline{e}^{k+1}_{ij}
\end{equation}
with $M$ being the number of incoming edges to node $n_j$.

Finally, the next node values $\underline{h}^{k+1}_i$ are computed using a (dense) neural network layer $f^k_\psi(\cdot)$.
This can be formulated as
\begin{equation}
\underline{h}^{k+1}_{j} = f_\psi^k([\underline{e}^{k+1}_{agg, j}, \underline{h}^k_{j}])
\end{equation}
for the $j$-th node.
These three steps are performed in every layer with each layer having (dense) network layers $f_\psi^l(\cdot)$ and $f_\chi^l(\cdot)$.
In this work, we do not use a global update as proposed in \cite{Battaglia}.

\subsection{Reinforcement Learning}
Reinforcement learning (RL) is a solution class for Markov decision processes (MDPs).
Contrary to dynamic programming or Monte-Carlo methods, RL does not require knowledge of the environment's dynamics but only learns from experiences.
RL solution methods can be divided into value-based, policy-based, and actor-critic (AC) approaches.

AC methods have an actor that learns a policy $\pi(s)$ and a critic that learns a state-value function $V(s)$ with $s$ being the state.
Most AC methods use a stochastic policy $\pi(s)$ that has a distributional output-layer.
In this work, we use an actor network that outputs a normal distribution $\mathcal{N}(\mu, \sigma)$ with $\mu$ being the mean and $\sigma$ the standard deviation.
The state-value function can either be learned using temporal differences (TD) learning or Monte-Carlo methods \cite{Sutton2018}.
We utilize TD learning to learn the state-value function $V(s)$.
The policy $\pi_\phi(a|s)$ and the state-value function $V_\xi(s)$ are approximated using deep neural networks and, therefore, are parameterized by the network weights $\phi$ and $\xi$.

The policy update for the actor using TD learning is defined as
\begin{equation}
  \nabla J =  (r_t + \gamma V_\xi(s_{t+1}) - V_\xi(s_t)) \nabla \log \pi_\phi(a_t | s_t)
	\label{eq:log_eq}
\end{equation}
with $r_t$ being the reward, $s$ the state, $a_t$ the action and $V_\xi(s)$ the approximated state-value function at time $t$.
Equation \ref{eq:log_eq} increases the (log-) likelihood of an action if the expected return is large and decreases it otherwise.
In this work, we use the proximal policy optimization (PPO) actor-critic algorithm that shows \emph{state-of-the-art} performance in various applications \cite{Dillmann, Morita}.
The PPO uses a surrogate objective function that additionally clips Equation \ref{eq:log_eq} to avoid large gradients in the update step.

The work is further structured as follows:
In the next section, we will provide related work of RL, GNNs, and the combination of both. 
In Section \ref{sec:approach}, we will go into detail of how we apply RL and GNNs for decision-making in autonomous driving.
And finally, we will provide experiments, results and give a conclusion.

\section{Related Work}
\label{sec:related_work}
In this section, we will outline and discuss related work of graph neural networks (GNNs), actor-critic (AC) reinforcement learning and the combination of both.

\subsection{Reinforcement Learning}
Reinforcement learning (RL) solution methods can be categorized in three categories: value-based, policy-based, and actor-critic methods \cite{Sutton2018}.
Of these three categories, the combination of value-based and policy-based RL in the form of AC methods have shown \emph{state-of-the-art} performance in continuous and dynamic control problems \cite{Schulman2015, Schulman, Lillicrap2015, Abdolmaleki2018}.

The trust region policy optimization (TRPO) algorithm restricts the updated policy to be close to the old policy \cite{Schulman2015}.
This is achieved by using the Kullback-Leibler (KL) divergence as a constraint in the optimization of the policy network.
They additionally prove that the TRPO method exhibits monotonically improving policies.
Since it is computationally expensive to calculate the KL divergence in every policy update, the proximal policy optimization (PPO) has been introduced \cite{Schulman}.
Instead of using the KL divergence, the PPO uses a clipped surrogate objective function.
The optimization of the clipped surrogate objective function can be done using unconstrained optimization and is less computationally expensive.

The soft actor-critic (SAC) method introduces an additional entropy term that is maximized \cite{Haarnoja2018}.
The SAC method, thus, tries to find a policy that is as random as possible but still maximizes the expected return.
As shown in their work, this yields the advantage that the agent keeps trying to reach different goals and does not focus (too early) on a single goal.
However, the SAC method uses action-value functions $Q(s, a)$ instead of a state-value function $V(s)$.
As this would introduce additional complexity combing GNNs with the SAC algorithm, we use the PPO algorithm in this work.
When using conventional neural networks the maximum number of vehicles and their order has to be specified.
Therefore, either a maximum number of vehicles or hand-crafted features are often utilized.
\citet{isele2018navigating} discretize an intersection using a grid world and use this as input for the neural network.
However, some information is lost due to discretization errors.

\citet{huegle2019dynamic} propose to use deep sets (DS) in order to mitigate the changing number and order of vehicles.
DS is invariant to the number and order of the inputs.
However, DS does not contain any relational information and the network has to learn these implicitly.
Contrary to that, GNNs can directly operate on graphs and utilize contained relational information.

Graph neural networks and reinforcement learning have been used together in various applications.
\citet{WangLBF18} propose NerveNet where GNNs are used instead of conventional deep neural networks.
By applying the same GNN to each joint, such as in the humanoid walker the GNN learns to generalize better and to handle and control each of these joints.

GNNs have also been used to learn state-representations for deep reinforcement learning \cite{sanchez2018graph, khalil2017learning}.

\citet{Boedecker} propose a deep scenes architecture, that learns complex interaction-aware scene representations.
They show the deep scenes architecture using DS and GNNs.
They use the GNN in combination with a Q-learning algorithm that directly learns the policy.

Contrary to their work, we use AC methods to learn continuous and stochastic policies for the ego vehicle.
Furthermore, we conduct studies on the robustness of conventional and graph neural networks.
Contrary to Q-learning, the PPO algorithm is an on-policy method that can lead to a more efficient exploration of the configuration space.
The risk of becoming stuck in local optima can be lowered by e.g.\ additionally optimizing the expected entropy as the SAC algorithm does.

\section{Approach}
\label{sec:approach}
This section describes how the graph is built, outlines the architecture of the actor- and critic-networks, and explains how graph neural networks (GNNs) and actor-critic (AC) reinforcement learning are combined for decision-making in autonomous driving.

In the semantic environment are $M$ vehicles with each having a state $\underline{s}_i$ that e.g.\ contains the velocity and the vehicle angle $\theta$.
A `GraphObserver' observes the environment from the ego vehicle's perspective and generates a graph with nodes $n_i \in N$ that are connected by edges $e_{ij} \in E$ with $i$ and $j$ being the node indices.
Vehicles that are within a threshold radius $r_{near}$ are included in the graph generation.
All vehicles within this radius are connected to their $n$-nearest vehicles.

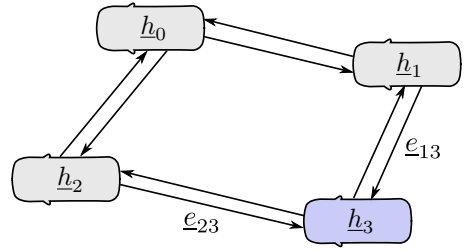
\begin{figure}[t!]
	\vspace{2mm}
	\centering
	\def\svgwidth{8.0cm}
\begingroup%
  \makeatletter%
  \providecommand\color[2][]{%
    \errmessage{(Inkscape) Color is used for the text in Inkscape, but the package 'color.sty' is not loaded}%
    \renewcommand\color[2][]{}%
  }%
  \providecommand\transparent[1]{%
    \errmessage{(Inkscape) Transparency is used (non-zero) for the text in Inkscape, but the package 'transparent.sty' is not loaded}%
    \renewcommand\transparent[1]{}%
  }%
  \providecommand\rotatebox[2]{#2}%
  \newcommand*\fsize{\dimexpr\f@size pt\relax}%
  \newcommand*\lineheight[1]{\fontsize{\fsize}{#1\fsize}\selectfont}%
  \ifx\svgwidth\undefined%
    \setlength{\unitlength}{245.66908949bp}%
    \ifx\svgscale\undefined%
      \relax%
    \else%
      \setlength{\unitlength}{\unitlength * \real{\svgscale}}%
    \fi%
  \else%
    \setlength{\unitlength}{\svgwidth}%
  \fi%
  \global\let\svgwidth\undefined%
  \global\let\svgscale\undefined%
  \makeatother%
  \begin{picture}(1,0.52467115)%
    \lineheight{1}%
    \setlength\tabcolsep{0pt}%
    \put(0,0){\includegraphics[width=\unitlength,page=1]{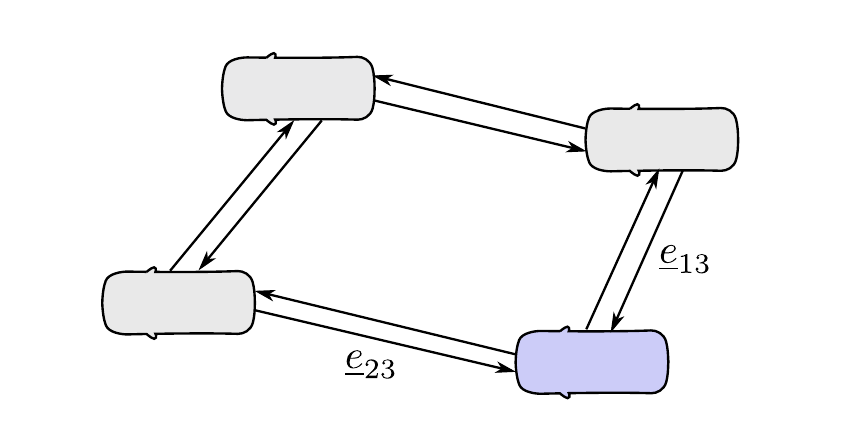}}%
    \put(0.67730289,0.09074688){\makebox(0,0)[lt]{\lineheight{1.25}\smash{\begin{tabular}[t]{l}$\underline{h}_3$\end{tabular}}}}%
    \put(0.75812926,0.35014061){\makebox(0,0)[lt]{\lineheight{1.25}\smash{\begin{tabular}[t]{l}$\underline{h}_1$\end{tabular}}}}%
    \put(0.33286817,0.4121165){\makebox(0,0)[lt]{\lineheight{1.25}\smash{\begin{tabular}[t]{l}$\underline{h}_0$\end{tabular}}}}%
    \put(0.19452579,0.15964831){\makebox(0,0)[lt]{\lineheight{1.25}\smash{\begin{tabular}[t]{l}$\underline{h}_2$\end{tabular}}}}%
  \end{picture}%
\endgroup%

	\caption{Directed graph $G=(N, E)$ in which the vehicles are connected to their two nearest neighbors.
           Each node $n_i$ and edge $e_{ij}$ with $i$ and $j$ being the node indices has vectorial information -- e.g.\ the nodes storing intrinsic and the edges relational information.}
	\label{fig:graph_structure}
\end{figure}

The node value $\underline{h}_i$ of the node $n_i$ contains intrinsic information of the $i$-th vehicle in form of a tuple $\langle x, y, v_x, v_y\rangle$ with $x, y$ being the cartesian coordinates and $v$ the velocity components.
The edge value $\underline{h}_{ij}$ of the edge $\underline{e}_{ij}$ between node $n_i$ and $n_j$ contains relational information in form of relative distances composed of two components $\langle d_x, d_y \rangle$.
The structure of the graph $G$ is depicted in Figure \ref{fig:graph_structure}.

A further component -- the `Evaluator' -- determines the reward signal $r_t$ for each time $t$ and whether an episode is terminal.
The reward signal $r_t$ is composed of scalar values that rate the safety and comfort of the learned policies.
It can be expressed as
\begin{equation}
r_t = r_{col} + r_{goal, reached} + r_{goal, dist} + r_{vel} + r_{act}
\label{eq:reward_signal}
\end{equation}
rating the collisions, reaching the goal, the distance to the goal, deviating from the desired velocity and penalizing large control commands, respectively.
The goal is reached once the ego vehicle has reached a defined state configuration -- a pre-defined range of $x, y$, $v$ and $\theta$.
The reward signal $r_t$ is weighted to avoid collisions and to create comfortable driving behaviors.

As outlined in Section \ref{sec:introduction}, we use the GNN approach proposed by \citet{Battaglia} with slight modifications.
Contrary to their work, we do not make use of global node features.
The GNN directly operates on graphs that are structured as in Figure \ref{fig:graph_structure}.

\begin{figure}[t!]
	\vspace{2mm}
	\centering
	\def\svgwidth{8.0cm}
	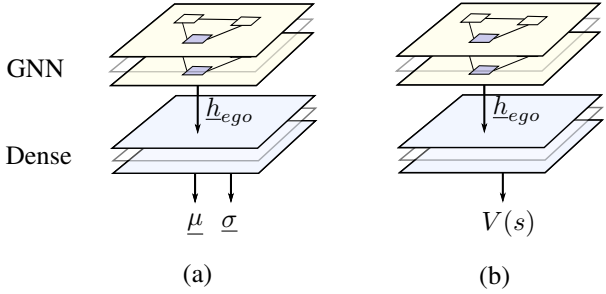
	\caption{Neural network structures used in this work:
           (a) shows the architecture of the actor having GNN layers and dense layers that output normal distribution parameters -- the means $\underline{\mu}$ and the standard deviations $\underline{\sigma}$ of the policy $\pi_\theta(\cdot)$.
           (b) depicts the architecture of the critic network that predicts the expected return $V_\xi(s)$. The critic network is also composed of GNN and dense layers.}
	\label{fig:network_graph_mlp}
\end{figure}

In the proposed approach, the actor network of the PPO directly takes the graph $G$ as the input and maps it to output distributions of the control commands -- the steering-rate $\delta$ and the acceleration $a$.
A normal distribution $\mathcal{N}(\mu, \sigma)$ for each of the control commands is used.
By default, the GNN outputs a value for each vehicle in the graph.
As we are only interested in controlling the ego vehicle, we only use the node value of the ego vehicle $\underline{h}_{ego}$.
This node value is then passed to a projection network that generates a distribution for the steering-rate $\delta$ and the acceleration $a$.
The projection network has dense layers and takes the node value of the ego vehicle $\underline{h}_{ego}$ as input.
The projection network builds distributions using the means $[\mu_0, \dots, \mu_k]$ and the standard deviations $[\sigma_0, \dots, \sigma_k]$ of each control command with $k$ being the number of control commands.
In order to limit the control commands, we additionally use a $\tanh(\cdot)$ squashing layer to restrain the network outputs to a certain range.
During training, the distributions are sampled to explore the environment and during application (exploitation) the mean $\mu$ is used.
This network represents the policy $\pi_\theta(\cdot)$ of the PPO algorithm with $\theta$ being the neural network parameters.
The architecture of the GNN actor network is depicted in Figure \ref{fig:network_graph_mlp} (a).

The critic network of the PPO has a similar architecture to the actor network.
It also directly operates on the graph $G$ and selects the node value of the ego vehicle $\underline{h}_{ego}$ in the output layer of the GNN.
The value of the ego vehicle node $\underline{h}_{ego}$ is then passed into a dense layer and mapped to a scalar value that approximates the expected return.
Using temporal difference learning, the state-value function $V_\xi(s)$ with $s$ being the state and $\xi$ being the neural network parameters is learned.

The node value of the ego vehicle $\underline{h}_{ego}$ in the GNN has always the same vectorial size regardless of the number and order of the vehicles in the semantic environment.
Unlike conventional neural networks, the maximum number of vehicles for the observation does not have to be pre-defined and fixed when using GNNs.
The only additional hyper-parameters that are introduced are added in the graph generation -- the threshold radius $r_{near}$  and with how many vehicles each vehicle is connected.
However, information of not directly connected vehicles can still be propagated through the graph due to the convolutional characteristics of GNNs.

In the next section, we conduct experiments, evaluate the novel approach, and compare it to using conventional neural networks.

\section{Experiments and Results}
\label{sec:experiments}
In this section, we conduct experiments and present results of our approach using graph neural networks (GNNs) as function approximator within the proximal policy optimization (PPO) algorithm.
We compare the proposed approach with using conventional deep neural networks for the actor and critic network.
As an evaluation scenario, we chose a highway lane-changing scenario with a varying number of vehicles.
Additionally, we conduct ablation studies that evaluate the generalization capabilities of both approaches.

\begin{figure}[t!]
	\vspace{2mm}
	\centering
	\def\svgwidth{8cm}
	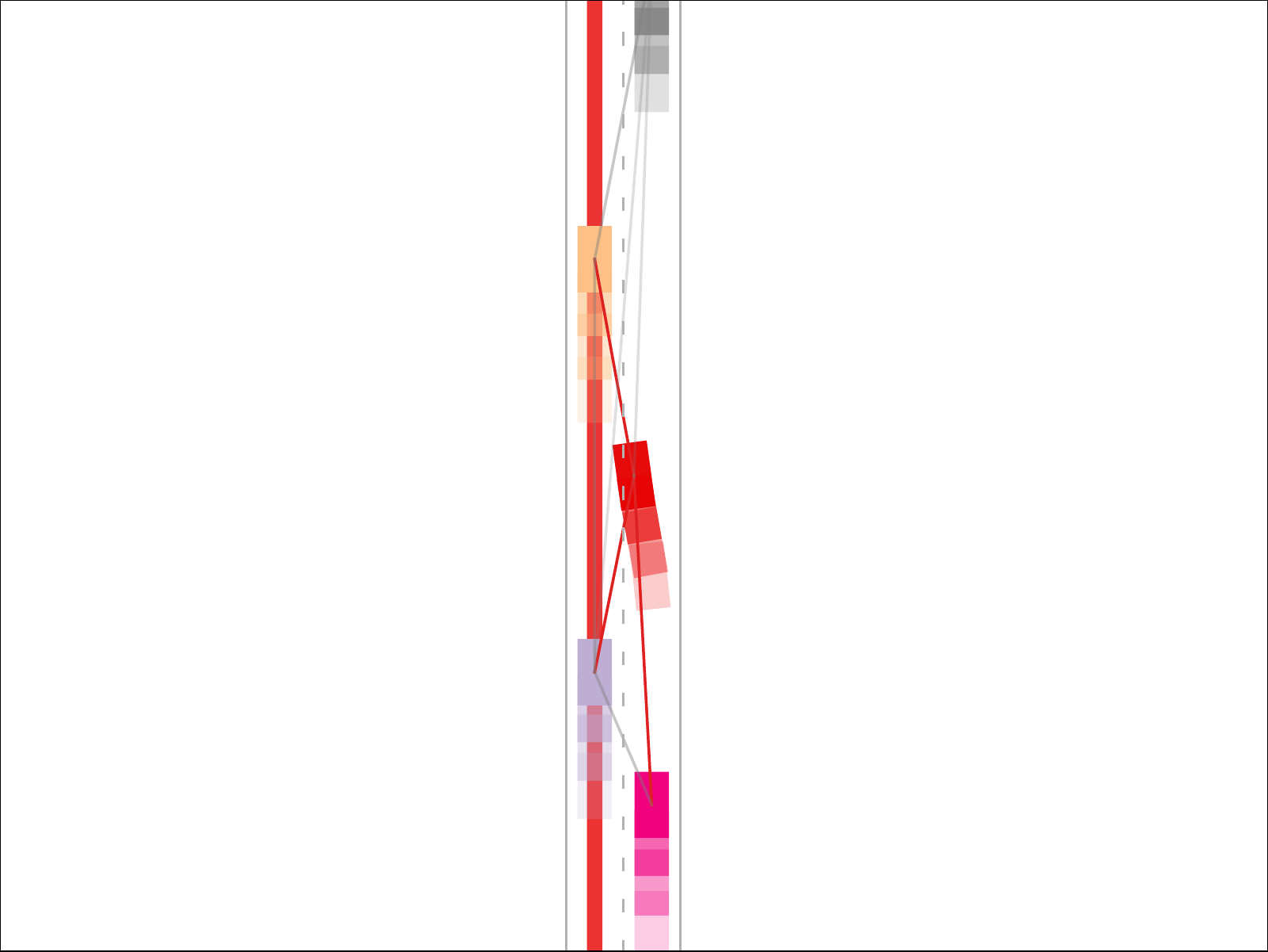
	\caption{The vehicle depicted in red is the ego vehicle and its goal is depicted as a polygonal shape on the left lane in the same color.
					 The first-order edges of the ego vehicle are depicted in red and all other edges in the graph are shown in gray.
           The other vehicles are controlled by the intelligent driver model.}
	\label{fig:xy_plot}
\end{figure}

All simulations are run using the BARK simulator \cite{bark}.
The ego vehicle is uniformly positioned on the right lane and its `StateLimitsGoal' goal definition is positioned on the left lane.
Thus, the ego vehicle tries to change the lane in order to achieve its goal.
All vehicles besides the ego vehicle are controlled by the intelligent driver model (IDM) parametrized as stated in \cite{treiber2013traffic}.
These vehicles follow their initial lane and do not change lanes.
The vehicles -- including the ego vehicle -- are assigned an initial velocity that is sampled in a range of $[10m/s, 15m/s]$.
The scenario used for training and validation is depicted in Figure \ref{fig:xy_plot}.

The reward signal $r_{t}$ for time $t$ is a weighted sum of the following terms:
\begin{itemize}
	\item $r_{goal, dist}$ squared $L2$ distance to the state-goal,
	\item $r_{vel}$ squared deviation to the desired velocity,
	\item $r_{act}$ squared and normalilzed control commands of the ego vehicle,
	\item $r_{col} = -1$ collision with the road boundaries or other vehicles and, 
	\item $r_{goal, reached} = +1$ if the agent reaches its goal.
\end{itemize}

The reward signal is additionally weighted to prioritize safety over comfort -- $r_{col}$ is weighted more prominently than the other terms.
An episode is counted as terminal once the defined goal has been reached or a collision with the ego vehicle has occurred.
The `StateLimitsGoal` definition checks whether the vehicle angle $\theta$, the distance to the center-line $r_{c}$, and the desired speed $v_{des}$ are within a pre-defined range.

As we focus on higher-level and interactive behavior generation, we neglect forces such as friction and use a simple kinematic single track vehicle model as used in \cite{HartK18}. 
This vehicle model has been parameterized with a wheel-base of $2.7m$.
To avoid large integration errors (especially of the IDM) we choose a simulation step-time $\Delta t=0.2s$.

The actor and critic networks are optimized using the Adam optimizer with a learning rate $lr = 3e-4$.
In this work, the actor and critic networks have identical structures.
For the GNN we choose a layer depth of $l=3$ with each node and edge layer having $80$ neurons and for the conventional neural network (NN) we use dense layers having $[512, 256, 26]$ neurons.
All layers in this work use ReLU activation functions to mitigate the vanishing gradients problems of neural networks.
In the next section, we will compare the performance of both networks used in the PPO algorithm.

\subsection{Conventional vs. Graph Neural Networks}
In this section, we compare the performance of conventional neural networks (NNs) with graph neural networks (GNNs).
The number of vehicles varies in every scenario as the positions of the vehicles are uniformly sampled on the road.
At most there are $12$ vehicles in the scenario given the used scenario configuration.

\begin{table}[h]
	\centering
	\begin{tabular}{ c | c | c | c }
		Scenario & Network & Success-rate [\%] & Collision-rate [\%] \\
		\hline
		Nominal   & NN & 81.0 \% & 18.2 \% \\
		\cline{2-4}
		  		    & GNN & 81.6 \% & 11.1 \% \\
		\hline
		Ablation & NN & 70.0 \% & 28.2 \% \\
		\cline{2-4}
		  				& GNN & 80.4 \% & 12.6 \% \\
	\end{tabular}
	\caption{Results of using NNs and GNNs for the lane-changing scenario.
					 In the `Nominal' scenario, the training and the evaluation are performed in the same scenario.
					 In the `Ablation' study, the observations of the vehicles are perturbed by changing the order of the observations.
					 All approaches have been evaluated using $100$ scenarios.}
	\label{tab:table_gnn_ffnn}
\end{table}

Both configurations have been trained using the same hyper-parameters.
For the NN we use a `NearestAgentsObserver' that senses the three nearest vehicles, sorts these by distance to the ego vehicle, and concatenates their states into a 1D vector.
The ego vehicle's state is added as the first state to this 1D vector. 
The GNN uses the before-described `GraphObserver' that connects each vehicle to its nearest neighboring vehicles.
We use $n=3$ for the number of nearest vehicles and a threshold radius $r_{near}=50m$.
These are the only additional hyper-parameters that are required when using GNNs.

Table \ref{tab:table_gnn_ffnn} shows the success- and the collision-rates for both approaches.
Both -- the conventional and the graph neural network -- are capable of learning the lane-changing scenario well.
In the `Nominal' case both networks have almost the same success-rate.
However, additional to a higher success-rate, the GNN also has a lower collision-rate.
The relatively high collision-rate can be justified that we do not check the scenarios for feasibility.
This means that some of the scenarios might not solvable due to the steering-rate and the acceleration of the ego vehicle being limited.
Thus, also optimal solutions might still cause collisions.

\subsection{Ablation Studies}
We conduct studies on how well conventional neural networks (NN) and graph neural networks (GNNs) cope with a changing order of vehicle observations.
We use the trained agents that have been used for evaluation in Table \ref{tab:table_gnn_ffnn} and the scenario shown in Figure \ref{fig:xy_plot}.
The scenarios have a varying number of vehicles and once a vehicle reaches the end of its driving corridor it is removed from the environment.
This results in a varying number of vehicles in the scenario.
Additionally, we now add noise to the sensed distances to other vehicles.
This has the effect that the observations are being changed in both observers.
The changing order and number of the vehicles models sensing inaccuracies that are persistent in the real-world due to e.g.\ sensor errors and faults.

In the `NearestAgentsObserver' adding noise to the distance results in perturbing the concatenated observation vector as the order of the vehicles is changed.
For the `GraphObserver' the perturbed distances change the edge connections of the graph resulting in the vehicles not only being connected to their nearest vehicles.

The results of the ablation study are shown in Table \ref{tab:table_gnn_ffnn}.
The GNN shows a higher robustness towards the order of the vehicles.
The success-rate remained high and the collisoin-rate only increased slightly.
Whereas in the conventional neural network, the success-rate decreased and the collision-rate increased significantly.
Due to several layers and convolution characteristics of GNNs information can be propagated over several nodes in the network -- e.g.\ from vehicles that the vehicle is not directly connected to.
This shows a higher invariance of GNNs towards perturbations in the observation space.
Additionally, as the ego vehicle's state is always in the first position when using the `NearestAgentsObserver', the NN still roughly can infer which actions to take regardless of the other vehicles.

\section{Conclusion}
In this work, we showed the feasibility of graph neural networks for actor-critic reinforcement learning used in semantic environments.
Both -- conventional and graph neural networks -- were able to learn the lane-changing scenario well.
We compared the performance of GNNs to conventional neural networks and showed that GNNs are more robust and invariant to the number and order of vehicles.

We outlined advantages that make using GNNs more favorable than using conventional neural networks.
GNNs do not require a fixed maximum number of inputs and are invariant towards the order of the vehicles in the environment.
They use relational information that is available in the graph and do not implicitly have to infer these relations.
Another advantage of GNNs is, that they make it possible to split intrinsic and extrinsic information.
For example, the nodes can store the vehicle information and the edges the relational information between two vehicles. 

We also performed ablation studies in which we changed the order of the vehicles.
This showed that GNNs generalize better and are more invariant to the order of the vehicles compared to conventional neural networks.
The success- and collision-rate of the GNN only dropped slightly whereas more significant changes are seen when using conventional neural networks.

In further work, additional edges to boundaries, traffic entities (such as traffic lights), and goals could be added and investigated.
This could drive the approach towards a more universal behavior generation approach.

\printbibliography

\end{document}